\documentclass[11pt,a4paper]{article}
\usepackage[hyperref]{acl2021}
\usepackage{times}
\usepackage{latexsym}

\usepackage{microtype}
\usepackage{graphicx}
\usepackage{multirow}
\usepackage{array}
\aclfinalcopy
\usepackage{amsmath} 
\usepackage{amssymb} 
\usepackage{amsfonts} 
\usepackage{booktabs}
\usepackage{arydshln}
\usepackage{tcolorbox}
\usepackage{verbatim}
\usepackage{xcolor}
\usepackage[loadshadowlibrary]{todonotes}
\usepackage{tcolorbox}
\usepackage{setspace}

\title{Retrieval-Augmented Semantic Parsing:\\ Improving Generalization with Lexical Knowledge}

\author{Xiao Zhang \\
  University of Groningen \\
  \texttt{xiao.zhang@rug.nl} \\
  \And
  Qianru Meng \\
  Leiden University \\
  \texttt{q.r.meng@}\\
\texttt{liacs.leidenuniv.nl} \\
  \And
  Johan Bos\\
  University of Groningen \\
  \texttt{johan.bos@rug.nl}
}

\date{}

\begin{document}
\maketitle
\begin{abstract}

Open-domain semantic parsing remains a challenging task, as neural models often rely on heuristics and struggle to handle unseen concepts. In this paper, we investigate the potential of large language models (LLMs) for this task and introduce Retrieval-Augmented Semantic Parsing (RASP), a simple yet effective approach that integrates external lexical knowledge into the parsing process. Our experiments not only show that LLMs outperform previous encoder-decoder baselines for semantic parsing, but that RASP further enhances their ability to predict unseen concepts, nearly doubling the performance of previous models on out-of-distribution concepts. These findings highlight the promise of leveraging large language models and retrieval mechanisms for robust and open-domain semantic parsing.

\end{abstract}

\section{Introduction\label{sec:intro}}

%
Open-domain semantic parsing involves mapping natural language text to formal meaning representations, capturing the concepts, relations between them, and the contexts in which they appear \cite{oepen-lonning-2006-discriminant, hajic-etal-2012-announcing, Banarescu2013AbstractMR, gmb, martinez-lorenzo-etal-2022}. Such meaning representations are applied in many downstream applications---ranging from database querying to embodied question answering---where parsers must handle a vast array of concepts that may not appear in the training data. While neural encoder-decoder architectures have shown impressive performance in semantic parsing tasks, their reliance on training distributions constrains their ability to generalize, especially to out-of-distribution (OOD) concepts.

Most existing semantic parsers struggle to interpret the symbols, such as rare senses, often defaulting the unseen words to the most frequent meaning encountered during training. As a result, they fail to adapt to novel linguistic phenomena and remain limited to fixed patterns. Recent work \cite{zhang-etal-2025-neural} have attempted to mitigate these limitations by encoding concept representations symbolically, forcing models to learn underlying structural knowledge from resources like WordNet \cite{Fellbaum-1998-wordnet}. However, these approaches require substantial preprocessing and intricate encodings that can be difficult for models to fully exploit.

In our work, instead, we explore the potential of large language models, powerful decoder-only architectures with strong in-context learning capabilities and extensive pretraining, to enhance the ability of semantic parsers to generalize. We pose two central research questions:

\begin{itemize}
    \item \textbf{Do large language models outperform traditional encoder-decoder architectures in semantic parsing?} Decoder-only architectures are known to be more scalable and to internalize broader knowledge, potentially leading to stronger generalization and learning abilities. Assessing their performance in semantic parsing tasks can help reveal the architectural advantages of these decoder-only models.
    \item \textbf{How can these large language models be leveraged to improve the generation of out-of-distribution concepts?} Beyond simple architecture comparisons, we investigate whether LLMs can be guided to handle concepts more flexibly, using their ability to interpret and integrate external information.
\end{itemize}

In Section~\ref{sec:background} we provide background on the semantic formalism of our choice, earlier approach to semantic parsing, and the challenge of an important task, word sense disambiguation. Then we propose \textbf{Retrieval-Augmented Semantic Parsing} (RASP) in Section~\ref{sec:method}, a technique that integrates a retrieval mechanism into parsing. RASP leverages external lexical knowledge in the input, enabling the model to dynamically access and interpret relevant concept information. By incorporating this retrieval step (Section~\ref{sec:exp}), we relax the reliance on lemma-based mappings and allow the model to adapt more naturally to unseen words or senses. Our results show that this approach nearly doubles the performance on predicting OOD concepts compared to previous methods, demonstrating a substantial advancement in handling challenging open-domain data (Section~\ref{sec:results}).

\section{Background and Related Work\label{sec:background}}

\subsection{Discourse Representation Structure}

\begin{figure}
    \centering
    \includegraphics[width=\linewidth]{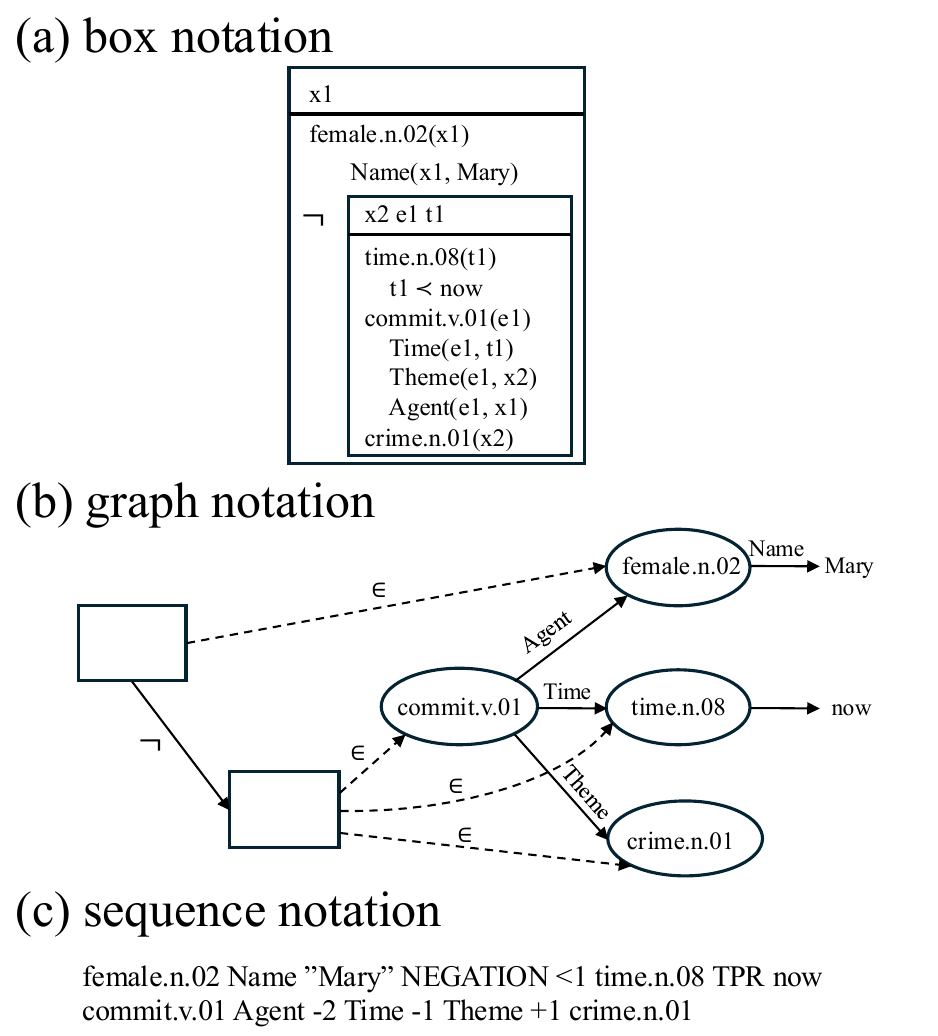}
    \caption{Three formats of Discourse Representation Structure (DRS) for  "Mary didn't commit a crime.": the box notation, a directed acyclic graph, and the sequence notation. These formalisms are mutually convertible without loss of information.}
    \label{fig:drs_example}
\end{figure}

Discourse Representation Theory \cite[DRT]{Kamp1993FromDT} is a semantic modeling framework. The core component of DRT is the Discourse Representation Structure (DRS), a formal representation that captures the meaning of a discourse, which captures the essence of the text and covers linguistic phenomena like anaphors and temporal expressions. Unlike many other formalisms such as Abstract Meaning Representation \cite[AMR]{Banarescu2013AbstractMR} used for large-scale semantic annotation efforts, DRS covers logical negation, quantification, and discourse relations. Moreover, DRS is equipped with complete word sense disambiguation, and offers a language-neutral meaning representation.  A Discourse Representation Structure (DRS) can be coded and visualised in various ways, which are all provided in Parallel Meaning Bank \cite{pmb}. In formal semantics they are often pictured in a human-readable box format. The clause notation was introduced to represent DRS in a sequential format suitable for machine learning models \cite{van-noord-etal-2018-exploring}. To further simplify DRS, \citet{Bos2023IWCS} proposed a variable-free format known as Sequence Box Notation (SBN). An example of the three different but logically equivalent formats is shown in Figure~\ref{fig:drs_example}. Recent trends in using seq2seq models have led to a preference for sequence notation, which is also the format used in this paper.

\subsection{Semantic Parsing}

Semantic parsing, as a traditional NLP task, remains essential in real-world applications, despite recent progress in natural language understanding shown by large language models. For instance, natural language front-end interfaces to databases require a mapping from text to structured data. Speech interactions with conversational agents that act in the real world (e.g., service robots) require situation-sensitive symbol grounding. Hence, advancing the development of more robust and general semantic parsers remains crucial.

Early approaches to semantic parsing primarily relied on rule-based systems \cite{woods1973progress,Hendrix1977DevelopingAN,templeton-burger-1983-problems}. The advent of neural methodologies, coupled with the availability of large semantically annotated datasets \cite{Banarescu2013AbstractMR, gmb, pmb}, marked a significant shift in semantic parsing techniques \cite{barzdins-gosko-2016-riga, noord-2017-neural, Bevilacqua2021OneST}. The introduction of pre-trained language models within the sequence-to-sequence framework further improved parsing performance \cite{ van-noord-etal-2018-exploring, van-noord-etal-2020-character, ozaki-etal-2020-hitachi, samuel-straka-2020-ufal, shou-lin-2021-incorporating, Bevilacqua2021OneST, Zhou2021AMRPW, martinez-lorenzo-etal-2022, zhang-etal-2024-gaining, liu-2024-model, liu-2024-soft, yang2024DRSparsing, amin2025semantic}. Furthermore, several studies introduced more pre-training tasks specifically designed for semantic parsing \cite{bai-etal-2022-graph, wang2023plmp}. With the rise of large language models, there has been considerable discussion about leveraging these models for semantic parsing, achieving notable results through techniques like prompting and chain-of-thought reasoning \cite{Roy2022BenchCLAMPAB, ettinger-etal-2023-expert, jin-etal-2024-analyzing}. However, there is currently no work that leverages the knowledge and understanding capabilities of large language models to address the generalization problem in semantic parsing.

\subsection{Word Sense Disambiguation}

The generalization problem introduced in the previous section can also be understood as word sense disambiguation (WSD) for out-of-distribution concepts, within the context of semantic parsing. For instance, consider the sentence "She had \pounds10,000 in the bank", with the target word "bank". In traditional WSD tasks, a predefined inventory of possible senses (e.g., 1.~sloping land; 2.~financial institution; 3.~a long ridge or pile; 4.~...) would be provided, and the WSD model would classify the word according to one of these senses \cite{navigli2009WSD, bevilacqua2021recent}. 

In semantic parsing, WSD can be seen as a subtask \cite{zhang-etal-2025-neural}, but it is more challenging because the parsing model must generate the correct sense directly without access to an explicitly provided sense inventory. However, traditional knowledge-based WSD offers a potential solution that inspires our approach: by retrieving and presenting all possible concepts as alternatives, we can explicitly provide external information to the model, thereby enhancing its generalization capability. 
As a consequence, this requires the model to be able to process long contexts, making the LLMs be the preferred choice, in particular retrieval augmented generation.

\subsection{Retrieval Augmented Generation}

Retrieval-Augmented Generation (RAG) is a hybrid approach that combines retrieval mechanisms with generative models to enhance the quality and accuracy of text generation tasks \cite{zhao2024rag, gao2024rag}. In RAG, a retrieval component first identifies relevant information from a large external knowledge base or corpus, which is then used as additional context for the generative model. This method allows the model to generate more informed and contextually accurate outputs, particularly in scenarios where the input data alone may not provide sufficient information. 

By integrating retrieved knowledge into the generative process, RAG effectively bridges the gap between retrieval and generative models, leading to improved performance in tasks such as question answering \cite{karpukhin-etal-2020-dense, patrick2020rag, Borgeaud2021ImprovingLM, guu2020retrieval, izacard-grave-2021-leveraging, petroni2020context}, commonsense reasoning \cite{liu2021kg, wan-etal-2024-evidence} and other downstream tasks \cite{patrick2020rag, Gautier2023Atlas, jiang-etal-2023-active, guo-etal-2023-prompt, cheng-etal-2023-uprise, li-etal-2023-structure}. While RAG was initially employed in a wide scope of applications, its popularity can be attributed to the advent of large language models and their strong capabilities. Consequently, we will concentrate on the application of RAG in the context of LLMs.

\section{Retrieval-Augmented Semantic Parsing\label{sec:method}}

\begin{table*}[htbp]
\small
\centering
\renewcommand{\arraystretch}{1.3}
\begin{tabular}{p{16mm} p{22mm} p{105mm}} 
\toprule
\textbf{Source Text}        & \multicolumn{2}{p{120mm}}{Mary went for birdwatching. She saw a harrier, a golden eagle, and a hobby.} \\ 
\midrule
\multirow{11}{*}{\centering \textbf{Concepts}} 
 & golden\_eagle.n.01: & large eagle of mountainous regions of ... having a golden-brown head and neck \\
 & birdwatch.v.01: & watch and study birds in their natural habitat \\
 & ... & ... \\
 & harrier.n.01: & a persistent attacker \\
 & harrier.n.02: & a hound that resembles a foxhound but is smaller \\
 & harrier.n.03: & hawks that hunt over meadows and marshes and prey on small terrestrial animals ... \\
  & ... & ... \\
 & hobby.n.01 & an auxiliary activity \\
 & hobby.n.02 & a child's plaything consisting of an imitation horse mounted on rockers ... \\
 & hobby.n.03 & small Old World falcon formerly trained and flown at small birds \\
\midrule
\multirow{2}{*}{\centering \textbf{Prompts}} 
 &  Normal prompt: & Text to parse: \{\emph{Source Text}\} \\
 &  RASP prompt: & Considering the concepts with glosses: \{\emph{Concepts}\}. Text to parse: \{\emph{Source Text}\} \\
\midrule
\textbf{Gold DRS}          & \multicolumn{2}{p{115mm}}{
\begin{tabular}[c]{@{}l@{}}
female.n.02 Name "Mary" time.n.08 TPR now birdwatch.v.01 Agent -2 Time -1 ELABORATION $<$1 \\ 
female.n.02 ANA -3 see.v.01 Experiencer -1 Time +1 Stimulus +3 time.n.08 TPR now harrier.n.03 \\ 
golden\_eagle.n.01 entity.n.01 Sub -2 Sub -1 Sub +1 hobby.n.03
\end{tabular}} \\ 
\bottomrule
\end{tabular}
\caption{An example illustrating the workflow of RASP. We omit some senses and words for the retrieved concepts to save space. The distinction between prompts for semantic parsing with and without RASP are shown in the \emph{Prompts} row. Some examples of complete prompts can be found in Appendix~\ref{app:prompt}.}
\label{tab:prompt}
\end{table*}

\begin{figure}
    \centering
    \includegraphics[width=\linewidth]{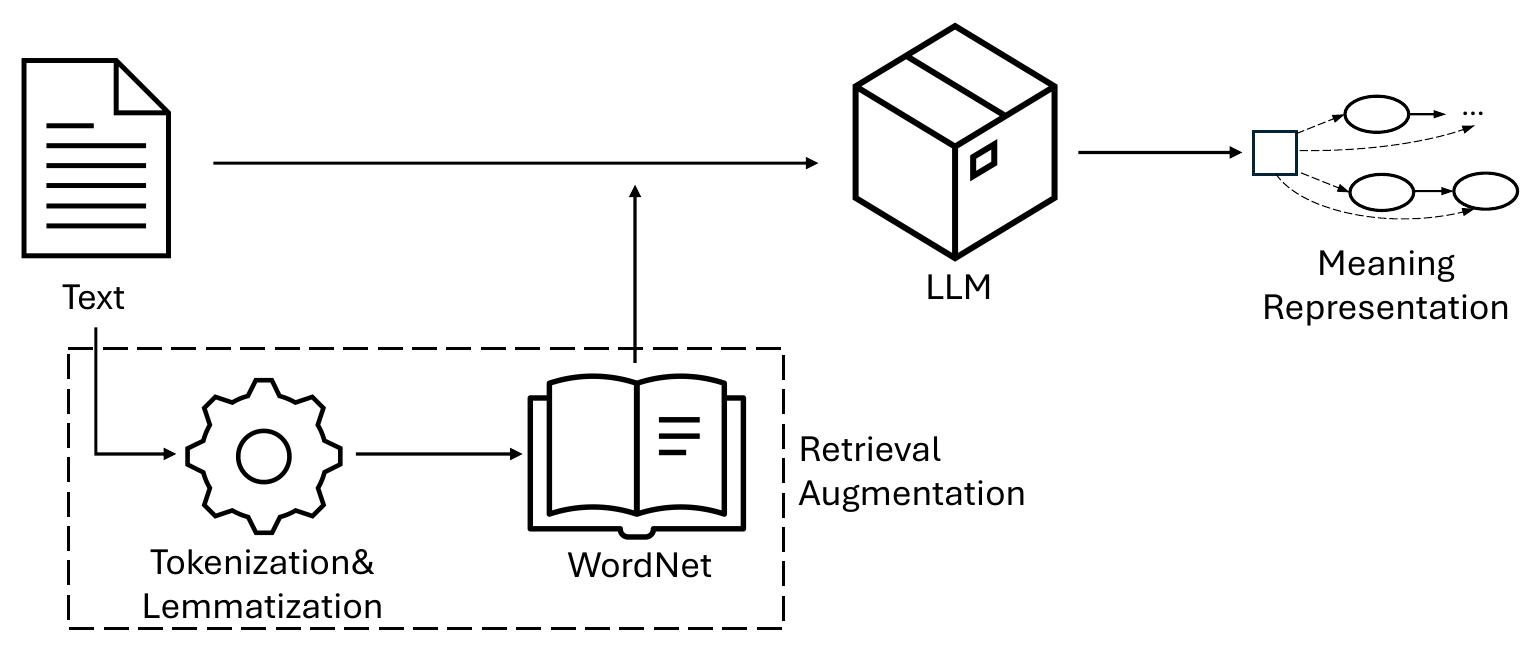}
    \caption{Global overview of RASP (Retrieval-Augmented Semantic Parsing). Both the training and testing phases adhere to this pipeline.}
    \label{fig:pipeline}
\end{figure}

We propose a new method that combines retrieval-augmented generation with semantic interpretation:  Retrieval-Augmented Semantic Parsing (RASP), a framework that is outlined in Figure~\ref{fig:pipeline}. It comprises two key components: retrieval and parsing.

Different from the Dense Passage Retrieval \cite{karpukhin-etal-2020-dense} method, which is commonly employed in question-answering tasks, our retrieval process is designed to be more straightforward and tailored to the needs of semantic parsing. The process begins with tokenizing and lemmatizing\footnote{https://www.nltk.org/api/nltk.stem.wordnet.html} the source text. Following these, we perform a search for relevant concept synsets in an external knowledge base, specifically WordNet. For example, in the sentence "Mary went for birdwatching. She saw a harrier, a golden eagle, and a hobby", the retrieval process would identify multiple synsets for "go", "birdwatch", "see", "harrier", "golden eagle" and "hobby", as illustrated in Table~\ref{tab:prompt}. Additionally, to ensure comprehensive coverage of multi-word expressions, which are critical in capturing the correct semantic meaning, we employ a hierarchical n-gram search strategy. This strategy involves sequential searches using 4-gram, 3-gram, 2-gram, and 1-gram patterns, thereby ensuring that no multi-word expressions (such as "golden eagle") are overlooked.


The parsing process for a decoder-only model\footnote{The models we use are all in decoder-only architecture, so we omit the discussion about encoder-decoder architecture.} is guided by the probability distribution of possible output sequences given an input sequence. The model generates an output sequence by predicting each token iteratively, based on the input text and previously generated tokens, as shown in (\ref{equ:de-only}).

{\small
\begin{align}
\small
p_{\text{decoder-only}}(o \mid x)
&= \prod_{i=1}^{n} p_{\theta}\!\left(o_i \mid x,\, o_{1:i-1}\right) \label{equ:de-only}
\end{align}
}

\noindent
Here, $x$ is the input text, $o_{1:i-1}$ represents the sequence generated so far, and $o_i$ is the token generated at the current step. $\theta$ refers to the model's parameters, and $p$ denotes the likelihood of generating output sequence $o$ given  input sequence $x$.


To enhance this process, retrieval and generation are integrated, leveraging external knowledge to inform output generation. Mathematically, the retrieval step introduces a probability, $p(o'|x)$, which models the likelihood of retrieving relevant concepts $o'$ based on $x$. This probability is combined multiplicatively with the generation probability, as shown in (\ref{equ:rap}). This combination ensures both components contribute meaningfully, with retrieval acting as a filter to guide the generation process toward relevant concepts.

{\small
\begin{align}
p_{\text{RASP}}(o \mid x)
&= p\!\left(o' \mid x\right)\; p_{\text{decoder-only}}\!\left(o \mid x,\, o'\right) \notag\\
&= p\!\left(o' \mid x\right)\; \prod_{i=1}^{n} p_{\theta}\!\left(o_i \mid x,\, o',\, o_{1:i-1}\right)
\end{align} \label{equ:rap}
}


By incorporating retrieved concepts, RASP goes beyond relying solely on the input sequence and training data, adding additional context to guide generation. For example, when handling words with multiple meanings, like "hobby," retrieved synsets help the model select the correct interpretation based on glosses and context. This integration sharpens the model's focus on relevant concepts, reducing the likelihood of generating incorrect or overly broad outputs, particularly for out-of-distribution concepts.


\section{Experiments\label{sec:exp}}

\subsection{Datasets}

We conduct our experiments on the Parallel Meaning Bank (PMB, version 5.1.0)\footnote{\url{https://pmb.let.rug.nl/releases}} \cite{pmb, zhang-etal-2024-gaining}. We first use the gold-standard English data of the PMB to evaluate the large language models and their retrieval-augmented version under in-distribution conditions.

To further assess the models’ ability to handle out-of-distribution (OOD) concepts, we adopt the challenge set proposed by \citet{zhang-etal-2025-neural}, which is also derived from the PMB. Neural semantic parsers often default to the first sense of unknown concepts--an approach that can lead to "lucky guesses" without truly understanding new words. The challenge set, consisting of 500 sentences, is deliberately designed to eliminate this shortcut. Each sentence includes at least one concept that does not appear in the training data and does not correspond to the first sense in the ontology. In total, the challenge set contains 410 unknown nouns, 128 verbs, and 65 modifiers (adjectives and adverbs). By evaluating on this set, we measure the true generalization capability of the models, testing whether they can correctly interpret novel concepts rather than relying on heuristic assignment.

\begin{table}[htbp]
\centering
\small
\renewcommand{\arraystretch}{1.2} 
\setlength{\tabcolsep}{10pt}
\begin{tabular}{cccc}
\toprule
\textbf{Train} & \textbf{Dev} & \textbf{Standard} & \textbf{Challenge} \\ \midrule
9,560          & 1,195        & 1,195              & 500               \\ \bottomrule
\end{tabular}
\caption{Dataset statistics for PMB 5.1.0, i.e., number of meaning representations for train, development and two test sets: standard and challenge.}
\label{tab:dataset_stats}
\end{table}

\subsection{Experiment Settings}

It is crucial to note that large language models, when used in zero-shot or few-shot scenarios, tend to perform poorly on the highly complex graph structures inherent in formal meaning representations such as DRS. Prior work \cite{ettinger-etal-2023-expert, zhang-etal-2025-neural} demonstrates that without fine-tuning, LLMs struggle to match the performance of models specifically optimized for these tasks. Therefore, in our experiments, we fine-tune all large language models. 

For RASP, we explore two retrieval-enhanced approaches: (1) Train+Test Retrieval: Incorporate retrieval-derived concepts both during training and inference, thereby familiarizing the model with external lexical knowledge throughout the entire learning process. (2) Test-Only Retrieval: Use retrieval only during inference, training the model on raw DRS structures without external lexical inputs. Our experiments show that the first approach consistently yields better performance. Thus, we focus our primary analysis on the first approach and provide results for the second approach in Appendix~\ref{app:experiments}.

Due to computational constraints, we select open-sourced LLMs with model sizes under 10B parameters, including phi3-4B, Mistral-7B, LLaMa3.1-3B, LLaMa3.2-8B, Gemma2-2B, Gemma2-9B, Qwen2.5-3B, and Qwen2.5-7B. These models strike a balance between state-of-the-art language understanding and manageable resource requirements. For fine-tuning, we employ Low-Rank Adaptation \cite[LoRA]{hu2022lora}, a parameter-efficient technique that introduces trainable low-rank matrices into the model’s layers, greatly reducing computational overhead.


We compare our results against several strong baselines, including BART, T5, byT5, TAX-parser \cite{zhang-etal-2024-gaining}, and AMS-Parser \cite{yang2024DRSparsing}, all of which were previously fine-tuned on PMB data. We exclude work conducted on earlier versions of PMB or using silver data. Additionally, we do not apply retrieval augmentation to these baseline models due to input length constraints, which limit their ability to incorporate external lexical sources efficiently.

We trained each model for 10 epochs, using a learning rate of $10^{-4}$, and fp16 precision. More information on the hyperparameters is provided in Appendix~\ref{app:settings}.

\begin{table*}[htbp]
\centering
\small
\begin{tabular}{l l l l l l l}
\toprule
\multirow{2}{*}{\textbf{Model}} & \multirow{2}{*}{\textbf{Size}} & \multirow{2}{*}{\textbf{Input}} & \multicolumn{3}{c}{\textbf{Graph-level}} & \textbf{Node-level} \\ 
\cmidrule(lr){4-6}
\cmidrule(lr){7-7}
                                &              &                                     & \textbf{Hard-SMatch$\uparrow$} & \textbf{Soft-SMatch$\uparrow$} & \textbf{IFR$\downarrow$} & \textbf{F score$\uparrow$}  \\ 
\midrule
BART-large                      & 400M        & Normal                              & 79.54 & 82.81 & 3.92 & 75.40      \\
T5-large                        & 770M        & Normal                              & 84.27 & 86.44 & 6.41 & 79.88      \\
byT5-large                      & 580M        & Normal                              & 87.41 & 89.43 & 4.78 & 84.75      \\  
AMS-Parser                      & --           & Normal                              & 87.08 & 89.15 & 0.00 & 85.00  \\
TAX-Parser                      & 580M           & Normal                              & 86.65 & 91.80 & 2.34 & 80.12    \\
\midrule
\multirow{2}{*}{Phi3}     & \multirow{2}{*}{4B}           & Normal                              & 85.74 & 87.92 & 4.94 (59) & 81.60      \\
                                &              & RASP                                & 85.96 (+0.3\%) & 88.13 (+0.2\%) & 4.80 (57) & 83.33 (+2.1\%) \\
                                \midrule
\multirow{2}{*}{Mistral}     & \multirow{2}{*}{7B}           & Normal                              & 89.95 & 92.48 & 2.00 (24) & 83.90      \\
                                &              & RASP                                & 90.95 (+1.1\%) & 93.33 (+0.9\%) & 1.58 (19) & 85.00 (+1.3\%) \\
                                \midrule
\multirow{4}{*}{Qwen2.5}       & \multirow{2}{*}{3B}           & Normal                              & 86.50 & 88.64 & 4.69 (56) & 82.60      \\
                                &              & RASP                                & 88.70 (+2.5\%) & 90.74 (+2.4\%) & 3.01 (36) & 83.90 (+1.6\%) \\\cdashline{2-7}
                                & \multirow{2}{*}{7B}           & Normal                              & 89.88 & 91.83 & 2.51 (30) & 84.50      \\
                                &              & RASP                                & 89.93 (+0.1\%) & 91.87 (+0.1\%) & 2.51 (30) & 85.50 (+1.2\%) \\\midrule
\multirow{4}{*}{LLama3}         & \multirow{2}{*}{3B}           & Normal                              & 87.30 & 90.01 & 3.34 (40) & 81.50      \\
                                &              & RASP                                & 87.76 (+0.5\%) & 90.51 (+0.6\%) & 3.01 (36) & 82.30 (+1.0\%) \\\cdashline{2-7}
                                & \multirow{2}{*}{8B}           & Normal                              & 89.92 & 92.46 & 2.09 (25) & 83.90      \\
                                &              & RASP                                & 90.65 (+0.8\%) & 93.10 (+0.7\%) & \textbf{1.50} (18) & 84.72 (+1.0\%) \\\midrule
\multirow{4}{*}{Gemma2}      & \multirow{2}{*}{2B}           & Normal                              & 89.20 & 91.08 & 3.01 (36) & 84.20      \\
                                &              & RASP                                & 89.30 (+0.1\%) & 91.23 (+0.2\%) & 3.10 (37) & 85.58 (+1.6\%)
                                \\\cdashline{2-7}
                                & \multirow{2}{*}{9B}           & Normal                              & 90.72 & 93.15 & 1.67 (20) & 84.67      \\
                                &              & RASP                                & \textbf{91.37} (+0.7\%) & \textbf{93.65} (+0.5\%) & 1.58 (19) & \textbf{86.11} (+1.7\%) \\
\bottomrule
\end{tabular}
\caption{Performance of baseline models, large language models (Normal) and their retrieval-augmented variants (RASP) on standard test, with percentage changes in parentheses. Size is the number of model's parameters (B: billion). IFR is Ill-Formed Rate and the number of ill-formed prediction are in parentheses. Note: AMS-Parser \cite{yang2024DRSparsing} performs well for IFR for it is a compositional neuro-symbolic system. TAX-Parser \cite{zhang-etal-2025-neural} is a neuro-symbolic system, trained with a novel encoded meaning representation.}
\label{tab:overall}
\end{table*}

\subsection{Evaluation Metrics}

We used SMATCH and its variants to evaluate the performance of the models. SMATCH \cite{cai-knight-2013-smatch}, referred to as Hard-SMatch, strictly matches concepts, where any discrepancy results in a non-match. In contrast, its variant, Soft-SMatch \cite{opitz-etal-2020-amr}, considers concept similarity when matching. Instead of adopting the approach of using word-embedding similarity, we applied the Wu-Palmer similarity \cite{wu-palmer-1994-verb}, as introduced by \citet{zhang-etal-2024-gaining}. Wu-Palmer similarity provides a precise measure of semantic similarity between concepts based on their positions within the WordNet taxonomy. Unlike embedding-based methods, it does not rely on external training and easily adapts to changes in WordNet's structure or content. The calculation is:

\begin{equation}
\small{
    \textrm{WuP} = 2 * \frac{depth(\textrm{LCS}(s_1, s_2))}{depth(s_1) + depth(s_2)}
    \label{equ:wps}}
\end{equation}

\noindent
where $s$ is the concept, LCS refers to the Least Common Subsumer of these concepts, and depth denotes the distance from the concept to the root of the taxonomy.

For the fine-grained evaluation on the challenge set, we applied the metric proposed by \citet{wang-etal-2023-discourse}, focusing specifically on concept-node matching scores. When evaluating the results on the challenge set, we directly calculated the Wu-Palmer similarity between the target concepts and the corresponding model-generated results.

\section{Results\label{sec:results}}


\subsection{Semantic Parsing on Standard Test}

Table~\ref{tab:overall} shows that large language models consistently surpass earlier encoder-decoder baselines, providing direct evidence for our first research question. While BART, T5, and byT5 achieve Hard-SMatch scores up to 87.41, several LLM-based models (e.g., Mistral-7B, Gemma2-9B) exceed 90.0 on the standard test set. This improvement is substantial, with the strongest baseline LLM reaches 90.72 on Hard-SMatch, outperforming the best encoder-decoder model (byT5) by a margin of 3.3 points. 

These higher scores are also reflected in Soft-SMatch and node-level F-scores, indicating that LLM-based models not only produce more structurally accurate meaning representations but also more reliably identify concept nodes. Additionally, Ill-Formed Rate (IFR) reductions suggest that these models generate fewer ill-structured outputs. In summary, these improvements highlight that large language models outperform previous encoder-decoder models.

Beyond confirming the advantages of LLMs, we also examine the impact of retrieval augmentation (RASP) on standard test results. Although the largest gains from retrieval are observed on the challenge set (as discussed in Section~\ref{subsec:challenge}), even here on the in-distribution standard test, RASP provides consistent performance improvements. Most LLMs show an increase of about 0.3\% to 2.5\% in Hard-SMatch and Soft-SMatch scores when using RASP. Furthermore, the Ill-Formed Rate (IFR) tends to decrease, and the node-level F-score improves by approximately 1.0\% to 2.1\%. These node-level gains suggest that RASP’s improvements stem largely from more accurate concept prediction. While these enhancements are moderate in the standard test scenario, they indicate that retrieval can enhance the model’s understanding of concept-level semantics.


\begin{table*}[h]
\centering
\small
\begin{tabular}{l l l l l l}
\toprule
\textbf{Model} & \textbf{Input} & \textbf{Noun} & \textbf{Verb} & \textbf{Modifiers}  & \textbf{Overall} \\ 
\midrule
BART-large     & Normal         & 26.11         & 37.34         & 46.88                     & 30.95             \\
T5-large       & Normal         & 25.48         & 35.21         & 41.28                  & 29.45             \\
byT5-large     & Normal         & 27.59         & 39.14         & 44.70                  & 32.13             \\
TAX-Parser     & Normal         & 42.15         & 31.58         & 43.27                 & 39.68             \\
\midrule
\multirow{2}{*}{phi3-4B} & Normal & 35.48         & 36.91         & 46.97              & 37.91             \\
                       & RASP   & 62.03 (+74.8\%) & 46.32 (+25.5\%) & 63.63 (+35.5\%)  & 58.28 (+53.7\%) \\
\midrule
\multirow{2}{*}{Mistral-7B} & Normal & 38.02         & 40.61         & 50.00                & 39.87             \\
                            & RASP   & 72.03 (+89.5\%) & 59.27 (+46.0\%) & 67.42 (+34.8\%)  & 68.44 (+71.7\%) \\\midrule
\multirow{2}{*}{Qwen2.5-7B} & Normal & 38.51         & 37.52         & 46.97                 & 39.12             \\
                            & RASP   & 66.77 (+73.4\%) & 56.95 (+51.8\%) & 64.39 (+37.1\%) & 64.12 (+63.8\%) \\\midrule
\multirow{2}{*}{LLama3.2-8B} & Normal & 37.06         & 34.79         & 47.73                   & 37.59             \\
                             & RASP   & 72.28 (+95.1\%) & 61.62 (+77.1\%) & 66.67 (+39.7\%)  & 69.86 (+85.9\%) \\
\midrule
\multirow{2}{*}{Gemma2-9B} & Normal & 39.68         & 45.01         & 55.30                   & 42.54             \\
                           & RASP   & \textbf{73.93} (+86.3\%) & \textbf{62.31} (+36.5\%) & \textbf{69.70} (+26.0\%)  & \textbf{70.41} (+65.6\%) \\
\bottomrule
\end{tabular}
\caption{Wu-Palmer similarities between unknown concepts and generated concepts across four parts of speech. For the sake of clarity, we exclude the smaller version of the same model.}
\label{tab:challenge}
\end{table*}





\subsection{Performance on the Challenge Set\label{subsec:challenge}}

Table~\ref{tab:challenge} provides the results on the challenge set, designed specifically to test the models' ability to predict out-of-distribution (OOD) concepts. Here, we report Wu-Palmer similarities for unknown nouns, verbs, and modifiers (adjectives and adverbs). We calculate the Wu-Palmer similarities between the target concepts (out-of-distribution concepts) and the generated concepts (see examples in Table~\ref{tab:case}).

Among the baselines, TAX-Parser \cite{zhang-etal-2025-neural} stands out, achieving an overall similarity score of 39.68. However, some Normal (non-RASP) large language models already exceed this performance on the challenge set. For example, Gemma2-9B (Normal) obtains an overall score of 42.54, indicating that LLMs can yield improvements, even without retrieval augmentation. When retrieval augmentation (RASP) is introduced, these large language models show substantial additional gains. For example, Gemma2-9B (RASP) achieves an overall similarity score of 70.41, compared to the best baseline’s 39.68---an increase of over 30 absolute points. These gains are particularly remarkable for noun concepts, with relative improvements of approximately 70\% to 95\%. Verbs show increases between about 25\% and 77\%, and modifiers improve by roughly 26\% to 43\%.

These results directly support our second research question regarding improving out-of-distribution generalization. While model scaling alone can yield moderate improvements, the integration of external lexical knowledge through retrieval allows LLMs to select more accurate concepts in OOD scenarios. In effect, RASP helps the models "look up" relevant information, enhancing their concept selection and producing more semantically appropriate results. In this case, retrieval-augmented LLMs not only outperform strong baselines like TAX-Parser but also set the state-of-the-art for OOD semantic parsing performance.



\subsection{Error Analysis on the Challenge Set}\label{sec:errors}

We selected a subset of the challenge set and manually checked how the best performing model---Gemma2-9B (Normal) and Gemma2-9B (RASP)---handle the out-of-distribution concepts. 


\begin{table*}[htbp]
\centering
\small
\setlength{\tabcolsep}{2pt}
\label{tab:cha_result}
\begin{tabular}{l>{\raggedleft\arraybackslash}p{2cm}>{\raggedleft\arraybackslash}p{2cm}>{\raggedleft\arraybackslash}p{2cm}}
\toprule
Input Text                          & Gold   & Normal & RASP \\ 
\midrule
He bought the painting for a \textbf{song} on a flea market.  &  song.n.05  & n.03 (0.22)  & n.05 (1.00) \\
The detective \textbf{planted} a bug in the suspect's office to gather evidence.         & plant.v.05 & v.02  (0.22) & v.05 (1.00) \\
Scientist examines the insect's \textbf{antennae}.    & antenna.n.03 & n.01 (0.24) & n.03 (1.00)\\
I’ve seen a short \textbf{extract} from the film.     & extract.n.02  & n.01 (0.25) & n.02 (1.00) \\
She prepared a three \textbf{course} meal.            & course.n.07 & n.03 (0.27) & n.07 (1.00) \\
The music student practiced the \textbf{fugue}.       & fugue.n.03  & n.02 (0.28)  & n.03 (1.00) \\
Johanna went birdwatching. She saw a harrier, a kite, and a \textbf{hobby}.  & hobby.n.03  & n.02 (0.38)  & n.03 (1.00)\\
A harrier is a \textbf{muscular} dog with a hard coat.  & muscular.a.02 & a.01 (0.50) & a.02 (1.00) \\
The hiker spotted an \textbf{adder} sunbathing on a rock.       & adder.n.03 & n.01 (0.50) & n.03 (1.00) \\
A tiny \textbf{wren} was hiding in the shrubs.        & wren.n.02 & n.01 (0.55) & n.02 (1.00) \\ 
\textbf{Hungarian} is a challenging language with 18 cases. & hungarian.n.02 & n.01 (0.11) & n.02 (1.00) \\
\midrule
The moon is \textbf{waxing}.                     & wax.v.03 & v.03 (1.00) & v.02 (0.75) \\
The function \textbf{ordered} the strings alphabetically. & order.v.05 & v.02 (0.17) & v.06 (0.75) \\
The elephant's trunk is an \textbf{extended} nose. & extended.a.03 & a.01 (0.50) & a.01 (0.50) \\
A \textbf{tripper} helps control the flow of materials on a conveyor.      & tripper.n.04 & n.02 (0.40) & n.02 (0.40) \\
We saw a \textbf{kite} gliding in the sky during the walking.   & kite.n.04 & n.03 (0.40) & n.03 (0.40) \\ 
The elegant \textbf{pen} glided gracefully across the tranquil lake.     & pen.n.05 & n.01 (0.36) & n.01 (0.36) \\
The immature sparrows are \textbf{feathering} already.    & feather.v.05 & v.03 (0.20) & v.02 (0.29) \\ 
The visitors can \textbf{observe} various species of ray in the aquarium. & observe.v.02 & v.01 (0.25) & v.01 (0.25) \\
She \textbf{hobbled} the horse. It freaked out.  & hobble.v.03 & v.01 (0.18) & v.02 (0.18) \\
The gardener noticed the \textbf{growth} on the rose after the rain. & growth.n.04 & n.01 (0.18) & n.01 (0.18) \\
The surge \textbf{alarmed} the town's residents.  & alarm.v.02 & v.01 (0.15) & v.01 (0.15) \\
\bottomrule
\end{tabular}
\caption{Twenty instances of the challenge set with content words with out-of-distribution concepts in bold face, and the concepts generated by the Gemma2-9B (Normal) and retrieval-augmented Gemma2-9B (RASP). The scores in brackets are the Wu-Palmer Similarity between the predicted concept and gold concept. }
\label{tab:case}
\end{table*}

We picked 22 instances, comprising 11 completely perfect predictions (WuP=1.00) and 11 imperfect predictions (WuP$<$1.00) made by RASP, as presented in Table~\ref{tab:case}. With respect to the perfect predictions, it is evident that the retrieval significantly enhances the model's ability to interpret most out-of-distribution concepts. For instance, in the text about birdwatching, the word "hobby" clearly refers to a species of bird. The model without RAG defaults to the most frequent sense number, predicts hobby.n.01 (an auxiliary activity). In contrast, retrieval provides the glosses of each sense related to the noun "hobby" and leads the model to pick hobby.n.03 (a falcon), by explicit lexical connections between "falcon" in the gloss of hobby.n.03 and the context provided by "birdwatching".

However, RASP makes imperfect predictions sometimes. We identified three possible causes: (a) similar glosses between WordNet concepts; (b) insufficient textual context; and (c) limitations in the model’s linguistic coverage. 

The verb "alarm" in Table~\ref{tab:case} is an instance of the similarity problem. The challenge arises because some of its senses have similar glosses, such as alarm.v.01 (fill with apprehension or alarm) and alarm.v.02 (warn or arouse to a sense of danger). Similar issues occur with the verbs "wax",  "order", "observe" and "hobble". Although glosses were carefully crafted by lexicographers, they don't always show a clear difference in meaning \cite{mihalcea2001automatic,navigli-2006-reducing}. 

In cases of insufficient textual context, such as with the noun "kite" in the sentence "We saw a kite gliding in the sky", the sense annotators chose kite.n.04 (a bird of prey). However, kite.n.03 (a plaything) could perhaps also be appropriate given the limited context provided by this sentence. Similar issues can be raised in the sentences with the noun "tripper" and the verb "feathering". 


The third cause can be attributed to the model's linguistic coverage. A case in point is "pen": the meanings of pen.n.01 (a writing implement) and pen.n.05 (a female swan) are quite different, but the latter is the correct one in the text "Jane saw two swans. The elegant pen glided gracefully across the tranquil lake". However, the model fails to distinguish them, likely because "pen" is rarely used to refer to "swan" in available corpora. As a result, the models may not have encountered this sense during training, making it challenging for them to predict a meaning they have not been exposed to.
In sum, while retrieval drastically improves concept prediction, there are still some difficulties that can pose challenges for the models.

\section{Conclusion}

This paper demonstrates that LLMs, even without retrieval augmentation, outperform previous encoder-decoder approaches in semantic parsing for Discourse Representation Structures, thereby answering our first research question in the affirmative, setting a new state of the art. We also show that our proposed Retrieval-Augmented Semantic Parsing (RASP) framework, which integrates external lexical knowledge, further enhances the performance of LLMs. Notably, RASP nearly doubles the accuracy on out-of-distribution concepts, which answers our second research question and confirms robust generalization ability of RASP in open-domain scenarios.
Our experiments show that by simply appending relevant information to the model input, the RASP approach offers a practical and intuitive approach that can be easily applied to other meaning representations used in natural language processing, such as AMR \cite{Banarescu2013AbstractMR} and BMR \cite{martinez-lorenzo-etal-2022}.

\newpage
\section{Limitations}

We think the limitations of this work mainly come from two aspects: the language models used in RASP and the retrieval source (i.e.,  WordNet).

The retrieval process is proven to provide more information and knowledge to the models. However, retrieval will significantly increase the input length of the model, making it (only) adoptable for the large language models with strong context understanding and long text processing capabilities. Therefore, the RASP framework cannot be directly used to improve previous parsers that rely on other methods, which is also why we only provided results of retrieval-augmented LLMs. 

Another limitation is  the retrieval source. Our implementation of RASP uses WordNet, so if a sense is not in WordNet, it will never be guessed. For example, "velvet scooter" (a bird) is not in WordNet, nor is Cobb salad (a dish). Hence, RASP will never make a perfect prediction for such cases.
Moreover, the glosses in WordNet, even though carefully crafted by lexicographers in most cases, are sometimes concise, lacking information to separate them from other senses. This makes it difficult for the models to accurately distinguish between different meanings (see Section~\ref{sec:errors}). For future work, the BabelNet, ConceptNet, or extended WordNet \cite{delmonte2012treebanks, Navigli2012BabelNetTA, delmonte2015logical, speer2017conceptnet} can be considered as a better choice for concept in meaning representations.

    

    

\bibliographystyle{acl_natbib}
\bibliography{acl2021}

\clearpage
\appendix

\section{Prompt \label{app:prompt}}
The following is a complete example of the prompts we use for the LLMs. Since the models we use are all instruction-based versions, the prompt is structured in a dialogue format.

\begingroup
\setlength{\parskip}{0pt}
\setlength{\parindent}{0pt}
\small

\begin{tcolorbox}[
  colback=white,
  colframe=gray,
  boxrule=0.5mm,
  arc=2mm,
  title=RASP Prompt,
  fonttitle=\bfseries\footnotesize,
  left=1mm, right=1mm, top=0.5mm, bottom=0.5mm,
  coltitle=black
]
\setstretch{0.9}
\footnotesize
\texttt{\textbf{user:}\\
Please parse the following text into Discourse Representation Structure, considering using the concepts based on the following glosses: \\
- birdwatch.v.01: watch and study birds in their natural habitat \\
- saw.n.01: a condensed but memorable saying embodying some important fact of experience that is taken as true by many people \\
- saw.n.02: hand tool having a toothed blade for cutting \\
- saw.n.03: a power tool for cutting wood \\
- saw.v.01: cut with a saw \\
- harrier.n.01: a persistent attacker \\
- harrier.n.02: a hound that resembles a foxhound but is smaller; used to hunt rabbits \\
- harrier.n.03: hawks that hunt over meadows and marshes and prey on small terrestrial animals \\
- kite.n.01: a bank check that has been fraudulently altered to increase its face value \\
- kite.n.02: a bank check drawn on insufficient funds at another bank in order to take advantage of the float \\
- kite.n.03: plaything consisting of a light frame covered with tissue paper; flown in wind at end of a string \\
- kite.n.04: any of several small graceful hawks of the family Accipitridae having long pointed wings and feeding on insects and small animals \\
- kite.v.01: increase the amount (of a check) fraudulently \\
- kite.v.02: get credit or money by using a bad check \\
- kite.v.03: soar or fly like a kite \\
- kite.v.04: fly a kite \\
- hobby.n.01: an auxiliary activity \\
- hobby.n.02: a child's plaything consisting of an imitation horse mounted on rockers; the child straddles it and pretends to ride \\
- hobby.n.03: small Old World falcon formerly trained and flown at small birds \\\\
Text to parse: Johanna went birdwatching. She saw a harrier, a kite, and a hobby.\\\\
\textbf{model:}\\
female.n.02 Name "Johanna" time.n.08 TPR now birdwatch.v.01 Agent -2 Time -1 ELABORATION <1\\
female.n.02 ANA -3 see.v.01 Experiencer -1 Time +1 Stimulus +3 time.n.08 TPR now\\
harrier.n.03 kite.n.04 entity.n.01 Sub -2 Sub -1 Sub +1 hobby.n.03}
\end{tcolorbox}

\vspace{0.5em} 

\begin{tcolorbox}[
  colback=white,
  colframe=gray,
  boxrule=0.5mm,
  arc=2mm,
  title=Normal Prompt,
  fonttitle=\bfseries\footnotesize,
  left=1mm, right=1mm, top=0.5mm, bottom=0.5mm,
  coltitle=black
]
\setstretch{0.9}
\footnotesize
\texttt{\textbf{user:}\\
Text to parse: Johanna went birdwatching. She saw a harrier, a kite, and a hobby.\\\\
\textbf{model:}\\
female.n.02 Name "Johanna" time.n.08 TPR now birdwatch.v.01 Agent -2 Time -1 ELABORATION <1 female.n.02 ANA -3 see.v.01 Experiencer -1 Time +1 Stimulus +3 time.n.08 TPR now harrier.n.03 kite.n.04 entity.n.01 Sub -2 Sub -1 Sub +1 hobby.n.03}
\end{tcolorbox}

\endgroup

\section{Experiment Settings\label{app:settings}}

Table~\ref{tab:config_overview} and ~\ref{tab:models} provide the basic details of the experiments and models.

\begin{table}[h!]
\centering
\small
\setlength{\tabcolsep}{8pt} 
\renewcommand{\arraystretch}{1.2} 
\begin{tabular}{@{}llll@{}}
\toprule
\textbf{Category}        & \textbf{Details}         & \textbf{Category}        & \textbf{Details} \\ \midrule
Stage                    & SFT/inference            & Precision                & fp16             \\
Fine-tuning              & LoRA                     & Batch Size               & 1                \\
Cutoff Length            & 1024                     & GPU Number               & 4                \\
Learning Rate            & $10^{-4}$                & GPU                     & H100              \\
Epochs                   & 10                       & lr scheduler            & cosine              \\ \bottomrule
\end{tabular}
\caption{Configurations for large language models Fine-Tuning and Inference.}
\label{tab:config_overview}
\end{table}

\begin{table}[h!]
\centering
\small
\setlength{\tabcolsep}{8pt} 
\renewcommand{\arraystretch}{1.2} 
\begin{tabular}{ll}
\toprule
\textbf{Model}           & \textbf{Details} \\ \midrule
BART-large               & facebook/bart-large \\
T5-large                 & google-t5/t5-large \\
byT5-large               & google/byt5-large \\
Phi3-4B                 & microsoft/Phi-3.5-mini-instruct \\
Qwen2.5-3B               & Qwen/Qwen2.5-3B-Instruct \\
Qwen2.5-7B               & Qwen/Qwen2.5-7B-Instruct \\
LLama3.2-3B              & meta-llama/Llama-3.2-3B-Instruct \\
LLama3.1-8B              & meta-llama/Llama-3.1-8B-Instruct \\
Gemma2-2B                & google/gemma-2-2b-it\\
Gemma2-9B                & google/gemma-2-9b-it\\
\bottomrule
\end{tabular}
\caption{Details of Models.}
\label{tab:models}
\end{table}

\section{Additional Experiments\label{app:experiments}}

We present the results of fine-tuning on Normal data and testing by RASP prompt, as shown in Tables~\ref{tab:extra_standard} and~\ref{tab:extra_challenge}. This approach involves providing retrieval information during inference but using only text-to-DRS data during training. From the results, it is evident that this training method adversely affects the model's performance, particularly on the standard test. We believe that fine-tuning reduces the models' ability of in-context learning, which limits the models from effectively utilizing the additional information provided by retrieval.

\begin{table*}[htbp]
\centering
\small
\begin{tabular}{l l l l l l l}
\toprule
\multirow{2}{*}{\textbf{Model}} & \multirow{2}{*}{\textbf{Size}} & \multirow{2}{*}{\textbf{Input}} & \multicolumn{3}{c}{\textbf{Graph-level}} & \textbf{Node-level} \\ 
\cmidrule(lr){4-6}
\cmidrule(lr){7-7}
&              &                                     & \textbf{Hard-SMatch$\uparrow$} & \textbf{Soft-SMatch$\uparrow$} & \textbf{IFR$\downarrow$} & \textbf{F score$\uparrow$}  \\ 
\midrule
\multirow{2}{*}{Phi3}     & \multirow{2}{*}{4B}           & Normal                              & 85.74 & 87.92 & 4.94 (59) & 81.60      \\
                                    &              & RASP                                & 66.78 (--22.1\%) & 70.88 (--19.4\%) & 14.9 (178) & 63.50 (--22.2\%) \\
\midrule
\multirow{2}{*}{Mistral}     & \multirow{2}{*}{7B}           & Normal                              & 89.95 & 92.48 & 2.00 (24) & 83.90      \\
                                    &              & RASP                                & 83.22 (--7.5\%) & 85.90 (--7.1\%) & 3.58 (43) & 80.10 (--4.5\%) \\\midrule
\multirow{4}{*}{Qwen2.5}       & \multirow{2}{*}{3B}           & Normal                              & 86.50 & 88.64 & 4.69 (56) & 82.60      \\
                                    &              & RASP                                & 84.32 (--2.5\%) & 87.44 (--1.4\%) & 5.00 (60) & 81.90 (--0.8\%) \\\cdashline{2-7}
                                    & \multirow{2}{*}{7B}           & Normal                              & 89.88 & 91.83 & 2.51 (30) & 84.50      \\
                                    &              & RASP                                & 86.23 (--4.1\%) & 90.78 (--1.1\%) & 2.57 (33) & 83.40 (--1.3\%) \\
\midrule
\multirow{4}{*}{LLama3}         & \multirow{2}{*}{3B}           & Normal                              & 87.30 & 90.01 & 3.34 (40) & 81.50      \\
                                    &              & RASP                                & 85.90 (--1.6\%) & 86.91 (--3.4\%) & 4.10 (49) & 77.59 (--4.8\%) \\\cdashline{2-7}
                                    & \multirow{2}{*}{8B}           & Normal                              & 89.92 & 92.46 & 2.09 (25) & 83.90      \\
                                    &              & RASP                                & 88.65 (--1.4\%) & 91.30 (--1.3\%) & 2.50 (30) & 82.11 (--2.1\%) \\
\midrule
\multirow{4}{*}{Gemma2}      & \multirow{2}{*}{2B}           & Normal                              & 89.20 & 91.08 & 3.01 (36) & 84.20      \\
                                    &              & RASP                                & 84.40 (--5.4\%) & 89.93 (--1.3\%) & 3.01 (36) & 80.11 (--4.9\%)
                                    \\\cdashline{2-7}
                                    & \multirow{2}{*}{9B}           & Normal                              & 90.72 & 93.15 & 1.67 (20) & 84.67      \\
                                    &              & RASP                                & 91.11 (+0.4\%) & 93.35 (+0.2\%) & 1.79 (21) & 83.10 (--1.9\%) \\
\bottomrule
\end{tabular}
\caption{Performance on standard test.}
\label{tab:extra_standard}
\end{table*}

\begin{table*}[h]
\centering
\small
\begin{tabular}{l l l l l l}
\toprule
\textbf{Model} & \textbf{Input} & \textbf{Noun} & \textbf{Verb} & \textbf{Modifiers} & \textbf{Overall} \\ 
\midrule
\multirow{2}{*}{phi3-4B} & Normal & 35.48 & 36.91 & 46.97 & 37.91 \\
                        & RASP   & 40.03 (+12.8\%) & 36.32 (--1.6\%) & 49.13 (+4.6\%) & 40.03 (+5.6\%) \\
\midrule
\multirow{2}{*}{Mistral-7B} & Normal & 38.02 & 40.61 & 50.00 & 39.87 \\
                            & RASP   & 40.90 (+7.6\%) & 49.27 (+21.3\%) & 50.00 (--0.0\%) & 43.61 (+9.4\%) \\\midrule
\multirow{2}{*}{Qwen2.5-7B} & Normal & 38.51 & 37.52 & 46.97 & 39.12 \\
                            & RASP   & 40.11 (+4.2\%) & 43.54 (+16.1\%) & 50.00 (+6.5\%) & 41.94 (+7.2\%) \\\midrule
\multirow{2}{*}{LLama3.2-8B} & Normal & 37.06 & 34.79 & 47.73 & 37.59 \\
                             & RASP   & 42.10 (+13.6\%) & 38.88 (+11.8\%) & 49.00 (+2.7\%) & 42.00 (+11.7\%) \\
\midrule
\multirow{2}{*}{Gemma2-9B} & Normal & 39.68 & 45.01 & 55.30 & 42.54 \\
                           & RASP   & \textbf{45.93} (+15.8\%) & \textbf{50.11} (+11.3\%) & \textbf{59.70} (+8.0\%) & \textbf{48.34} (+13.6\%) \\
\bottomrule
\end{tabular}
\caption{Performance on the challenge set.}
\label{tab:extra_challenge}
\end{table*}

\end{document}